\documentclass{tlp}
 
\usepackage{latexsym}
\usepackage{url}
\usepackage{aopmath}

\newcommand{\Mx}{\mathit{Max}}
\newcommand{\tn}{\mathit{T}^{\mathit{nd}}}
\newcommand{\td}{\mathit{T}^{\mathit{d}}}
\newcommand{\tp}{\mathit{T}}
\newcommand{\lar}{\leftarrow}
\newcommand{\At}{\mathit{At}}

\newcommand{\lm}{\mathit{lm}}
\newcommand{\hd}{\mathit{hd}}
\newcommand{\lparseNegfree}{$\n$-free lparse}

\newcommand{\bd}{\mathit{bd}}
\newcommand{\as}{\mathit{aset}}
\newcommand{\lst}{\mathit{lset}}
\newcommand{\hs}{\mathit{hset}}

\newcommand{\body}{\mathit{body}}

\newcommand{\n}{\mathbf{not}}

\newtheorem{definition}{Definition}
\newtheorem{example}{Example}

\begin{document}

\submitted{30 June 2004}
\revised{7 February 2006}
\accepted{23 August 2006}

\title[Theory and Practice of Logic Programming]
{Logic programs 
with monotone abstract constraint atoms\footnote{Parts of this paper
appeared earlier in conference papers \cite{mnt03,mt04}.}}

\pagerange{\pageref{firstpage}--\pageref{lastpage}}

\author[V.W. Marek, I. Niemel\"a and M. Truszczy\'nski]
{Victor W. Marek\\
Department of Computer Science, University of Kentucky, Lexington, KY 40506,
USA \\
\email{marek@cs.uky.edu}
\and 
Ilkka Niemel\"a\\
Department of Computer Science and Engineering\\
Helsinki University of Technology,\\
P.O.Box 5400, FI-02015 TKK, Finland \\
\email{Ilkka.Niemela@tkk.fi}
\and
Miros\l aw Truszczy\'nski\\
Department of Computer Science, University of Kentucky, Lexington, KY 40506,
USA \\
\email{mirek@cs.uky.edu}
}
\pagerange{\pageref{firstpage}--\pageref{lastpage}}

\maketitle

\begin{abstract}
We introduce and study logic programs whose clauses are built 
out of {\em
monotone constraint atoms}. We show that the operational concept of 
the one-step provability operator generalizes to programs with monotone
constraint atoms, but the generalization involves nondeterminism. Our 
main results demonstrate that our formalism is a common generalization 
of (1) normal logic programming with its semantics of models, supported 
models and stable models, (2) logic programming with weight atoms ({\em 
lparse} programs) with the semantics of stable models, as defined by 
Niemel\"a, Simons and Soininen, and (3) of disjunctive logic programming 
with the possible-model semantics of Sakama and Inoue.\\
To appear in {\em Theory and Practice of Logic Programming} (TPLP).
\end{abstract}

\section{Introduction}
\label{intro}

In this paper, we introduce and study logic programs whose clauses are 
built of generalized atoms expressing constraints on sets. 
We propose a generalization of normal logic programming to this extended
setting. Our generalization uses the assumption of the monotonicity of
constraints and it employs the nondeterminism in deriving ways to satisfy
constraints. In our approach the basic concepts, methods, semantics and
results of normal logic programming generalize to the proposed context.
Our work provides a theoretical framework to a recent extension of logic
programming with weight constraints (also known as {\em pseudo-boolean
constraints}) \cite{nss99,sns02}, and to an earlier 
formalism of disjunctive logic programs with the semantics of possible 
models \cite{si94}, but applies to a much broader class of programs. 

In the 1990s researchers demonstrated that normal logic programming with 
the stable-model semantics is an effective knowledge representation 
formalism. It provides solutions to problems arising in such contexts as 
planning, reasoning about 
actions, diagnosis and abduction, product 
configuration, and modeling and reasoning about preferences. Moreover, 
due to the emergence of fast methods to compute stable models 
\cite{ns97,cmodels,lz02,le-etal03}, the importance of the formalism 
increased significantly as it became possible to use it not only as a 
modeling language but also as a practical computational tool. 
The contributions of \cite{baral03,GelLeo02} provide a detailed 
discussion of the 
formalism and its applications. 

In the last few years, researchers proposed extensions of the language 
of normal logic programming with means to model constraints involving 
{\em aggregate} operations on sets. \cite{sns02} proposed a formalism 
integrating logic programming with {\em weight constraints}, known in
the SAT community as {\em pseudo-boolean} constraints, generalized 
the concept of stable models to this extended setting, and developed 
fast algorithms to compute them. \cite{dpb01,pdb04}, introduced a 
formalism allowing for more general aggregates. 
They extended to this 
new setting several semantics of normal logic programs, including the 
stable-model semantics and the well-founded semantics. 
A related recent 
work \cite{dlv-agg-03,flp04,cal:ijcai05}, incorporated aggregates into 
the 
formalism of {\em disjunctive} logic programs with the {\em answer-set 
semantics}. Yet another extension of normal logic programming has been
proposed in \cite{mr04} as set-based constraints. Such extensions are 
important as they simplify the task of 
modeling problem specifications, typically result in more direct and 
concise encodings, and often significantly improve the computational 
effectiveness of the formalism as a problem-solving tool.

Our goal is to propose an abstract formalism of logic programs extended 
with means to model constraints on sets, preserving as much as possible
analogies between our theory and {\em normal} logic programming. We 
introduce the notion of an {\em abstract constraint} and its linguistic 
counterpart --- an {\em abstract constraint atom}. We then use abstract 
constraint atoms as building blocks of program clauses in the same way
propositional atoms form clauses of normal logic programs. For the most
part, we restrict our attention to {\em monotone} constraints, as 
monotonicity is essential for preserving the notion of a logic program 
as a computational device. We show that basic concepts, techniques, and 
results of normal logic programming have direct generalizations 
for the 
class of programs built of monotone abstract constraints. What
distinguishes our work from other recent approaches to integrating
logic programming with aggregates is that we allow constraint atoms in
the heads of clauses, while formalisms proposed and studied in
\cite{dpb01,pdb04,dlv-agg-03,flp04,cal:ijcai05} do not\footnote{We note
though that recently \cite{spt06} also considered programs with constraints 
in the heads of rules.}.

In many respects the theory we built in this paper mirrors closely
an operator-based treatment of normal logic programs. There is, however,
a basic difference. Abstract constraint atoms are inherently
nondeterministic. They can be viewed as shorthands for certain
disjunctions and, in general, there are many ways to make an abstract
constraint atom true. This nondeterminism has a consequence. The
one-step provability operator, which generalizes the one-step
provability operator of \cite{vEK76} defined for normal programs, is no 
longer deterministic. It assigns to an interpretation $M$ a {\em set} 
$T(M)$ of interpretations. Each interpretation in the set $T(M)$ is 
regarded as possible and equally likely outcome of applying the 
operator to $M$.

The nondeterministic one-step provability operator is a key tool in our
work. It generalizes the one-step provability operator of \cite{vEK76}.
Thanks to close parallels between these two concepts, we are able to
reconstruct operator-based characterizations of models, supported
models, and the concept of a {\em bottom-up} computation for programs
with abstract constraints that generalize Horn programs. We
then extend to programs with abstract monotone constraints the 
definitions of the {\em Gelfond-Lifschitz} reduct and a stable model
\cite{gl88}. We also distinguish and discuss the class of {\em 
definite} programs (programs with clauses whose heads can be 
satisfied in one way only). For these programs the one-step provability 
operator becomes deterministic and the theory of normal logic 
programming extends to deterministic programs without {\em any} 
significant change. In particular, it follows that normal logic
programming with all its major 2-valued semantics can be viewed as
a special case of logic programming with monotone abstract constraints.

In addition, we show that programs with abstract constraints provide a 
formal account of a class of logic programs with weight (pseudo-boolean)
atoms introduced in \cite{sns02}. We call programs in the syntax 
proposed in that paper {\em lparse programs}. \cite{sns02} defined for
lparse programs the notion of a {\em stable model} and showed that
lparse programs generalize normal logic programming with the 
stable-model semantics of Gelfond and Lifschitz \cite{gl88}. However, 
the notion of the reduct underlying the definition of a stable model 
given in \cite{sns02} is different from that proposed in \cite{gl88} 
and the precise nature of the relationship between normal logic 
programs and lparse programs was not clear.

Our work explicates this relationship. On one hand, the formalism of 
programs with abstract constraints parallels normal logic programming. 
In particular, major concepts, results and techniques in normal logic 
programming have counterparts in the setting of programs with abstract
constraints. On the other hand, under some simple transformations,
lparse programs can be viewed as a special case of programs with 
abstract
constraints. Thus, through this connection, the theory of normal logic 
programming can be lifted to the setting of lparse programs 
leading, in 
particular, to new characterizations of stable models of lparse
programs.

Finally, we show that programs with monotone abstract constraints 
generalize the formalism of disjunctive logic programs with the semantics
of {\em possible} models \cite{si94}. In fact, as we point out, several 
ideas that are stated in abstract terms in our paper have their roots in 
\cite{si94}.

\section{Basic concepts, motivation, examples} 
\label{basic-concepts}

We consider a language determined by a fixed set $\At$ of {\em
propositional atoms}. 
An {\em abstract constraint} is a collection $C\subseteq {\cal P}(\At)$ 
(if $X$ is a set, by ${\cal P}(X)$ we denote the family of all subsets 
of $X$). 
We say that elements of $C$ {\em satisfy} the constraint $C$ or have 
the property $C$. An {\em abstract constraint atom} (or {\em ac-atom}, 
for short) is an expression $C(X)$, where $X \subseteq \At$ is {\em 
finite} and $C$ is an abstract constraint. An {\em 
ac-literal} is an expression of the form $C(X)$ or $\n(C(X))$,
where $C(X)$ is an ac-atom. We call $X$ the {\em atom set} of an ac-literal
$A$ of the form $C(X)$ or $\n(C(X))$ and denote it by $\as(A)$.

An intended meaning of an ac-atom $C(X)$ is to represent a requirement
on subsets of $X$ that they must satisfy the constraint $C$.
Formally, we interpret ac-atoms by means of 
propositional interpretations (truth assignments), which we represent 
as subsets of $\At$: an atom $p$ is {\em true} in an interpretation $M 
\subseteq\At$ if $p\in M$, otherwise, $p$ is {\em false} in $M$. An 
interpretation $M\subseteq\At$ {\em satisfies} an ac-atom $C(X)$,
written $M\models C(X)$, if $M\cap X\in C$ (that is, if the set of 
atoms in $X$ that are true in $M$ satisfies the constraint $C$). 
Otherwise, $M$ does not satisfy $C(X)$, written $M\not \models C(X)$. 
In that case, we also say that $M$ satisfies the {\em ac-literal} 
$\n(C(X))$ and write $M\models \n(C(X))$. An ac-atom $C(X)$ is {\em 
consistent} 
if there is an interpretation $M$ such that $M\models C(X)$ or,
equivalently, if $C$ contains at least one subset of $X$. We will now 
illustrate these concepts with several examples of common constraints.

\noindent
{\bf Pseudo-boolean constraints}. These constraints are also known as
{\em weight constraints}. Given a real number $w$ and a function 
$W$, assigning to each atom in $\At$ a real number (its {\em weight}),
a {\em pseudo-boolean} constraint $PB(w,W,\leq)$ imposes a restriction  
that ``the total weight of atoms in a set is at least $w$''. Formally, we
set $PB(w,W,\leq) = \{A \subseteq \At\colon w \leq \sum_{a\in A} W(a)\}$
(comparison relations $<$, $>$, $\geq$ give rise to other types of
weight constraints).\\
{\bf Cardinality constraints}. They are pseudo-boolean constraints
in which a bound $w$ is a non-negative integer and a weight of every
atom is 1. Throughout the paper, we use cardinality constraints to
illustrate concepts we study. To simplify the notation and to make it 
consistent with the notation 
used in \cite{sns02}, we write $kX$ to denote an 
ac-atom $C(X)$, where $C= PB(k,W,\leq)$ and $W(a)=1$ for every
$a\in \At$.\\
{\bf Product constraints.} They differ from weight constraints in
that they restrict the product of individual weights of atoms in
allowed sets, depending on the type of the comparison relation used.
Selecting the relation $\leq$ and assuming the same notation as before,
we express product constraints as abstract constraints of the form
$\Pi(w,W,\leq) = \{A\subseteq \At \colon w \leq \Pi_{a\in A} W(a)\}$.\\
{\bf Maximum constraints.} Given a weight function $W$ on the
set of atoms and a real bound $w$, the maximum constraint restricts
allowed sets of atoms to those with the maximum weight at least $w$.
Formally, we express them as abstract constraints of the form
$\Mx(w,W,\leq)= \{A\subseteq \At \colon w \leq \max\{W(a)\colon a\in
A\}\}$ (or its variants, depending on the comparison relation).\\
{\bf Even- and odd-cardinality constraints.} They impose a parity
requirement on the cardinality of allowed sets. Formally, we express
them as abstract constraints $E = \{A\subseteq \At \colon |A|\
\mbox{is even}\}$ and $O=\{A\subseteq \At\colon |A|\ \mbox{is odd}\}$.\\
{\bf Containment constraints.} Such constraints require that allowed
sets contain some prespecified configurations (subsets).
We capture them by abstract constraints $C({\cal A})$ that consist of
all subsets of $\At$ that contain at least one set from a
prespecified collection $\cal A$ of finite subsets of $\At$.

Each of these constraints determines associated ac-atoms. Let $\At=
\{p_1,p_2,\ldots\}$ and let us consider a weight function $W$ such
that for every integer $i\geq 1$, $W(p_i)=i$. The expression
$PB(6,W,\leq)$ is an example of a pseudo-boolean constraint. If we 
denote it by $C_1$, then $C_1(\{p_1,p_2,p_5,p_6\})$ is an example of 
a pseudo-boolean constraint atom. A set $M\subseteq\At$ satisfies 
$C_1(\{p_1,p_2,p_5,p_6\})$ if and only if the total weight of atoms 
in $M\cap \{p_1,p_2,p_5,p_6\}$ is at least $6$ (that is, if and only
if $M$ contains $p_6$, or $p_5$ together with at least one other
atom). Similarly, $\Mx(5,W,\leq)$ is an example of a maximum constraint
and, if we denote it by $C_2$, $C_2(\{p_2,p_4,p_6,p_8\})$ is a maximum
constraint atom that enforces the restriction on sets of atoms to
contain $p_6$ or $p_8$. An abstract constraint atom $E(\{p_1, p_7\})$ 
($E$ stands for the even-cardinality constraint) forces allowed sets of 
atoms to contain none or both of $p_1$ and $p_7$. All these constraint 
atoms are consistent. An atom $C_3(\{p_1,p_2,p_3\})$, where $C_3 = 
PB(7,w,\leq)$ is an example of an inconsistent 
constraint atom. No selection of 
atoms from $\{p_1,p_2,p_3\}$ satisfies it and, consequently, it has 
no models.

These examples demonstrate that abstract constraints and abstract 
constraint atoms express a broad range of common constraints. In this
paper, we show that abstract constraint atoms can be combined into 
logic program clauses to represent even more complex constraints, and 
that much of the theory of normal logic programs generalizes to the 
extended setting.

\section{Logic programs built of $\cal F$-atoms} \label{aclp}

Let $\cal F$ be a class of abstract constraints over $\At$. By an {\em 
$\cal F$-atom} we mean an abstract constraint
atom $A(X)$ such that $A\in {\cal F}$ 
and $X \subseteq \At$. 
An $\cal F$-{\em literal} (or simply, a {\em literal}, if the context is
clear) is an expression of the form 
$A(X)$ or $\n(A(X))$, where $A(X)$ is an $\cal F$-atom. 
An $\cal F$-{\em clause} is an expression 
\begin{equation}
\label{eq.r.1}
A(X) \leftarrow B_1(X_1),\ldots,B_m(X_m),
                \n(C_1(Y_1)), \ldots, \n(C_n(Y_n))
\end{equation}
where $A(X)$, $B_i(X_i)$ and $C_j(Y_j)$ are $\cal F$-atoms. An $\cal
F$-clause (\ref{eq.r.1}) is 
called a {\em constraint clause} if $A(X)$ is not 
consistent. An {\em $\cal F$-program} is a 
finite collection of $\cal F$-clauses\footnote{We note that the
assumption of the finiteness of programs is not essential. The entire 
theory of $\cal F$-programs extends to the case when we admit infinite 
programs. However, additional means of proof, such as Zorn Lemma, may 
be required in some arguments (for instance, in the argument for the 
existence of minimal models of $\cal F$-programs).}.

If $r$ is a clause of the form (\ref{eq.r.1}), $A(X)$ is the {\em head}
of $r$, denoted by $\hd(r)$, and $X$ is the {\em head set} of $r$, 
denoted by $\hs(r)$. We also call the conjunction of literals 
$B_1(X_1),\ldots,B_m(X_m), \n(C_1(Y_1)), \ldots, \n(C_n(Y_n))$, the {\em
body} of $r$ and denote it by $\bd(r)$. Occasionally, we use the same 
term to denote the {\em set} of all literals in the body of a clause. 
Finally, 
for an $\cal F$-program $P$, we define $\hs(P)$ to be the union of sets 
$\hs(r)$, for $r \in P$. 

An interpretation $M\subseteq \At$ {\em satisfies} a set (conjunction) 
$L$ of literals, if it satisfies every literal in $L$. We say that $M$
{\em satisfies} an $\cal F$-clause $r$ if $M$ satisfies the head of the 
clause whenever it satisfies the body of $r$. Finally, $M$ {\em 
satisfies} an $\cal F$-program $P$ if it satisfies all clauses in $P$. 
We write $M\models L$, $M\models r$ and $M\models P$ to denote these 
three types of the satisfaction relation. We will often write ``{\em is 
a model of}'' instead of ``satisfies''.
$\cal F$-programs that have models are called {\em consistent}. 

Clauses of normal logic programs are typically regarded as
computational devices: {\em assuming that preconditions of a
clause have been established, the clause provides a justification to
establish (compute) its head}. Crucial concepts behind formal accounts 
of that intuition are those of a Horn program, the corresponding {\em
bottom-up} computation, and a least Herbrand model, which defines the 
{\em result} of the computation. Computations and their results are 
well defined due to the {\em monotone} behavior of Horn programs. 

To extend normal logic programming to the class of programs with 
abstract constraint atoms, one needs a generalization of the class of 
Horn programs supporting an appropriate notion of a computation, with 
the results of computations playing the same role as that played by 
the least Herbrand model. In order to accomplish that, it is not enough 
simply to disallow the negation operator in the bodies of
$\cal F$-clauses. It is also necessary to restrict the class of 
constraints to those that are {\em monotone} (that is, intuitively, 
once true in an interpretation, they remain true in every 
superset of it). Without that assumption, the monotonicity of normal 
Horn programs does not generalize and there is no straightforward way
to define the concept of a computation. (We refer to \cite{vm05} for
a study of properties of monotone constraints.)

Formally, we say that an abstract constraint $C$ is {\em monotone}
if for every $A,A'\subseteq \At$, if $A\in C$ and $A \subseteq A'$ then
$A'\in C$ (in other words, monotone constraints are precisely
upward-closed families of subsets of $\At$). An abstract constraint
atom $C(X)$ is monotone if its constraint $C$ is monotone. 

\begin{proposition} 
\label{mono}
Let $C$ be an abstract monotone constraint over $\At$, $X\subseteq
\At$, and let $M,M' \subseteq \At$ be two interpretations. If $M
\models C(X)$ and $M \subseteq M'$, then $M'\models C(X)$.
\end{proposition}

We note that if all the individual weights used by a weight function
are non-negative, the corresponding pseudo-boolean constraints are 
monotone. The maximum constraints are monotone for {\em every} weight 
function. On the other hand, we note that some common constraints, for 
instance, even- and odd-cardinality constraints $E$ and $O$, are not 
monotone.

From now on we restrict our attention to constraints that are {\em 
monotone}. We will write a {\em monotone} $\cal F$-clause and a {\em 
monotone} $\cal F$-program to make it explicit that all constraints 
in $\cal F$ are monotone. 

An important consequence of the monotonicity assumption is that monotone
$\cal F$-programs without constraint clauses have models (and so, also 
minimal models).

\begin{proposition}
\label{model-exists}
Let $P$ be a monotone $\cal F$-program without constraint clauses. Then the set 
$\At$ of all atoms in the language is a model of $P$.
\end{proposition}
\begin{proof}
Let $r\in P$. Since the constraint $\hd(r)$ is consistent, there is a set of
atoms $M \subseteq \At$ such that $M\models \hd(r)$. By the
monotonicity of constraints in $\cal F$, $\At\models\hd(r)$. Thus,
$\At\models P$.
\end{proof}

Another important consequence of the monotonicity assumption is that
the concept of a Horn program has an obvious direct generalization. 

\begin{definition}
A monotone $\cal F$-program that contains no occurrences of the operator
$\n$ is a {\em Horn $\cal F$-program}. 
\ \ $\proofbox$
\end{definition}

Horn $\cal F$-programs defined in this way have many properties that
generalize well-known properties of {\em normal} Horn programs. We will
state and prove several of them later in the paper. 

\section{Nondeterministic one-step provability operator}
\label{nondet-op}

Following a fundamental idea underlying normal logic programming, we 
assign to $\cal F$-clauses a procedural interpretation, which views 
them as derivation clauses. 
In the discussion that follows we do not 
assume that constraints in $\cal F$ are monotone. 

Intuitively, if an $\cal F$-clause $r$ has its body satisfied by some 
set of atoms $M$, then $r$ provides {\em support} for deriving from $M$ 
any set of atoms $M'$ such that
\begin{enumerate}
\item $M'$ consists of  
some atoms from the headset of $r$ ($r$ provides
{\em no} grounds for deriving atoms that do not appear in its headset)
\item $M'$ satisfies the head of $r$ (since $r$ ``fires'', the
constraint imposed by its head must hold).
\end{enumerate}
Clearly, the process of deriving $M'$ from $M$ by means of $r$ is {\em
nondeterministic} in the sense that, in general, there are several sets
that are supported by $r$ and $M$. 

This interpretation of $\cal F$-clauses extends to $\cal F$-programs.
Given an $\cal F$-program $P$ and a set of atoms $M$, each clause $r\in 
P$ such that $M$ satisfies the body of $r$ provides a support for a 
subset of the head set of $r$. The union, say $M'$, of such sets --- 
each supported by {\em some} clause $r$, with $r$ ranging over those clauses 
in $P$ whose body is satisfied by $M$ --- can be viewed as ``derived'' 
from $M$ by means of $P$. In general, given $P$ and $M$, there may be 
several such derived sets. Thus, the notion of derivability associated 
with a program is nondeterministic, as in the case of individual 
clauses. 

We describe formally this intuition of derivability in terms of a 
nondeterministic one-step provability operator. Before we give a 
precise definition, we note that by a {\em nondeterministic operator} on 
a set $D$ we mean any function $f\colon D \rightarrow {\cal P}(D)$. One 
can view the set $f(d)$ as the collection of all possible outcomes of 
applying $f$ to $d$ one of which, if $f(d) \not=\emptyset$, can be 
selected nondeterministically as the {\em actual} outcome of $f$. We 
emphasize that we allow $f(d)$ to be empty, that is, nondeterministic 
operators are, in general, {\em partial} --- for some elements of the 
domain they do not assign any possible outcomes.

\begin{definition}
Let $\cal F$ be a class of constraints (not necessarily monotone). Let 
$P$ be an ${\cal
F}$-program and let $M\subseteq \At$.
\begin{enumerate}
\item A clause $r\in P$ is {\em $M$-applicable}, if $M\models\bd(r)$.
We denote by $P(M)$ the set of all
$M$-applicable clauses in $P$.
\item A set $M'$ is {\em nondeterministically one-step provable} from
$M$ by means of $P$, if $M' \subseteq \hs(P(M))$ and $M'\models \hd(r)$,
for every clause $r$ in $P(M)$.
\item The {\em nondeterministic one-step provability operator} $\tn_P$, 
is a function from ${\cal P}(\At)$ to ${\cal P}({\cal P}(\At))$ such 
that for every $M \subseteq \At$, $\tn_P(M)$ consists of all sets $M'$ 
that are nondeterministically one-step provable from $M$ by means of 
$P$.
\ \ $\proofbox$
\end{enumerate}
\end{definition}

Since an abstract constraint forming the head of an $\cal F$-clause may 
be inconsistent, there exist programs $P$ and interpretations $M
\subseteq\At$ such that $\tn_P(M)$ is empty.

The concepts introduced above have especially elegant properties for
{\em monotone} $\cal F$-programs. 
First, to illustrate them, let us consider a simple example involving
a program with cardinality constraints (cf. Section 
\ref{basic-concepts}). 
The program discussed in this example is not a Horn program.

\begin{example}
Let $P$ be a program with cardinality constraints consisting of 
the following clauses:
\begin{quote}
\noindent
$r_1=\ \ 2\{a\} \leftarrow 2\{b,d\}$\\
$r_2=\ \ 1\{b,c\} \leftarrow \n(1\{e\})$\\
$r_3=\ \ 1\{a,d\} \leftarrow 2\{b,c\}$
\end{quote}
We note that the cardinality atom in the head of the first clause
is inconsistent.

Let us consider a set $M=\{b,c,e\}$. Since $M\not\models 2\{b,d\}$,
$r_1$ is not $M$-applicable. Similarly, $M\not\models\n(1\{e\})$ and $r_2$
is not $M$-applicable, either. On the other hand, $M\models 2\{b,c\}$ 
and so, $r_3$ is $M$-applicable.

There are three subsets of $\{a,d\}$ that satisfy the constraint
$1\{a,d\}$ in the head of the clause $r_3$: $\{a\}$, $\{d\}$ and $\{a,
d\}$. Thus, each of these sets is nondeterministically one-step provable
from $M$ and, consequently,
\[
\tn_P(M) = \{\{a\}, \{d\}, \{a,d\}\}\mbox{.}
\]

We also note that if $|M|=1$ and $e\notin M$ then $r_2$ is the only 
$M$-applicable clause in $P$. For such sets $M$, $\tn_P (M) = \{\{b\},
\{c\},\{b,c\}\}$. 
On the other hand, if $M$ contains both $b$ and $d$, then $r_1$ is 
$M$-applicable and since the head of $r_1$ is inconsistent, $\tn_P(M)=
\emptyset$ (no set is nondeterministically one-step provable from such
a set $M$).
\ \ $\proofbox$
\end{example}

The example shows, in particular, that it may be the case that $\tn_P(M)
=\emptyset$. If, however, $P$ is a {\em monotone} $\cal F$-program without 
constraint clauses, then it is never the case. 

\begin{proposition}\label{ok}
Let $P$ be a monotone ${\cal F}$-program without constraint clauses. 
For every $M\subseteq \At$, $\hs(P(M))\in \tn_P(M)$. In particular, 
$\tn_P(M)\not=\emptyset$.
\end{proposition}
\begin{proof} 
Let us consider $r\in P(M)$. Then, $\hs(P(M))\cap \hs(r)=\hs(r)$. Since 
$\hd(r)$ is consistent, it follows by the monotonicity of constraints in
$\cal F$ that $\hs(r)\models \hd(r)$. Thus, $\hs(P(M)) \models \hd(r)$ 
and, consequently, $\hs(P(M))\in \tn_P(M)$.
\end{proof}

The operator $\tn_P$ plays a fundamental role in our research. It allows
us to formalize the procedural interpretation of $\cal F$-clauses and 
identify several classes of models. 

Our first result characterizes models of monotone $\cal F$-programs. 
Models of a normal logic program $P$ are prefixpoints of the one-step 
provability operator $\tp_P$ \cite{vEK76}. This characterization lifts 
to the class of monotone $\cal F$-programs, with the operator $\tn_P$ 
replacing $\tp_P$.

\begin{theorem}
\label{model}
Let $P$ be a monotone $\cal F$-program and let $M\subseteq \At$. The 
set $M$ is a model of $P$ if and only if there is $M'\in \tn_P(M)$ 
such that $M'\subseteq M$.
\end{theorem}
\begin{proof} 
Let $M$ be a model of $P$ and $M'=M\cap \hs(P(M))$. Let $r\in P(M)$. 
Since $M$ is a model of $r$, $M\models \hd(r)$. Clearly, $\hs(r)
\subseteq \hs(P(M))$. Thus, $M\cap \hs(r) = M' \cap \hs(r)$ and,
consequently, $M'\models \hd(r)$. It follows that $M'\in \tn_P(M)$.
Since $M'\subseteq M$, the assertion follows.

Conversely, let us assume that there is $M'\in \tn_P(M)$ such that 
$M'\subseteq M$. Let $r\in P$ be a clause such that $M\models \bd(r)$. 
Since $M'\in \tn_P(M)$, $M'\models \hd(r)$. We recall that the
constraint involved in $\hd(r)$ is monotone (as we consider only monotone 
constraints). Thus, by Proposition \ref{mono}, $M\models \hd(r)$, 
as well. It follows that $M$ is a model of every clause in $P$ and, 
consequently, of $P$. 
\end{proof}

\section{Supported models of ${\cal F}$-programs}
\label{supported}

For a set $M$ of atoms, we say that $M$-applicable clauses in an $\cal
F$-program $P$ provide {\em support} to atoms in the heads of these 
clauses. In general, a model $M$ of an $\cal F$-program may contain 
elements that have no support in a program and $M$ itself, that is, 
cannot be derived from $M$ by means of clauses in the program. 

\begin{example}
Let $P$ be a program with cardinality constraints 
consisting of a single clause:
\[
1\{p,q\} \lar \n(1\{q\}),
\]
where $p$ and $q$ are two different atoms. Let $M_1=\{q\}$. Clearly,
$M_1$ is a model of $P$. However, $M_1$ has no support in $P$ and
itself. Indeed, $\tn_P(M_1)=\{\emptyset\}$ and so, $P$ and $M_1$ do not
provide support for any atom. Similarly, another model of $P$, the set
$M_2=\{p,s\}$, where $s \in \At$ is an atom different from $p$ and $q$,
has no support in $P$ and itself. We have $\tn_P(M_2) = \{\{p\},\{q\},
\{p,q\}\}$ and so, $p$ has support in $P$ and $M_2$, but $s$ does not.
Finally, the set $M_3=\{p\}$, which is also a model of $P$, {\em has
support} in $P$ and itself. Indeed, $\tn_P(M_3)=\{\{p\},\{q\},\{p,q\}\}$
and there is a way to derive $M_3$ from $P$ and $M_3$. 
\ \ $\proofbox$
\end{example}

For $M$ to be a model of $P$, $M$ must satisfy the heads of all 
applicable clauses. To this end, $M$ needs to contain some of the
atoms appearing in the headsets of these clauses (atoms with support in $M$
and $P$) and, possibly, also some atoms that do not have such support.
Models that contain {\em only} atoms with support form an important 
class of models generalizing the class of supported models for normal 
logic programs \cite{cl78,ap90}.

\begin{definition}\label{supp}
Let $\cal F$ be a class of constraints (not necessarily monotone)
and let $P$ be an $\cal F$-program. A set of atoms $M$ is a {\em 
supported model} of $P$ if $M$ is a model of $P$ and $M\subseteq
\hs(P(M))$.
\ \ $\proofbox$
\end{definition}

Supported models have the following characterization generalizing
a characterization of supported models of normal logic programs as
fixpoints of the van Emden-Kowalski operator (the characterizing
condition is commonly used as a definition of a {\em fixpoint} of
a nondeterministic operator).

\begin{theorem}
\label{ch-sup}
Let $\cal F$ be a class of constraints (not necessarily monotone). Let 
$P$ be an $\cal F$-program. A set $M\subseteq\At$ is a supported model 
of $P$ if and only if $M \in \tn_P(M)$.
\end{theorem}
\begin{proof}
If $M$ is a supported model of $P$ then it is a model of $P$ (by the
definition). Moreover, $M\subseteq \hs(P(M))$. Thus, $M \in \tn_P(M)$.
Conversely, if $M \in \tn_P(M)$, then $M\subseteq \hs(P(M))$ and $M
\models \hd(r)$, for every $r\in P(M)$. Thus, $M\models r$, for every
$r\in P(M)$. If $r\in P\setminus P(M)$, then $M\not\models \bd(r)$ and
so, $M\models r$. Thus, $M$ is a model of $P$. Since $M \in \tn_P(M)$
also implies $M\subseteq \hs(P(M))$, $M$ is a supported model of $P$.
\end{proof}

In Section \ref{nlp-mca} we show that the use of the term {\em
supported} for the class of models defined in this section is not a
misnomer; supported models of $\cal F$-programs generalize supported
models of normal logic programs. 

\section{Horn ${\cal F}$-programs}
\label{horn}

For the concepts of the one-step provability and supported models we 
did not need a restriction to monotone constraints. To properly
generalize the notion of a stable model, however, this restriction
is essential. Thus, from this point on, we will consider only 
monotone $\cal F$-programs.

First, we will study Horn $\cal F$-programs (we recall that the notion
of a Horn $\cal F$-program assumes that $\cal F$ consists of monotone 
constraints only) viewing them as 
representations of certain nondeterministic computational processes. 
We will later use the results of this section to extend to the class
of $\cal F$-programs the concept of a stable model.

\begin{definition}
Let $P$ be a Horn
$\cal F$-program. A {\em $P$-computation} is a sequence $(X_n)_{n
=0,1,\ldots}$ such that $X_0=\emptyset$ and, for every non-negative
integer $n$:
\begin{enumerate}
\item $X_n\subseteq X_{n+1}$, and
\item $X_{n+1}\in \tn_P(X_{n})$.
\end{enumerate}
Given a computation $t=(X_n)_{n=0,1,\ldots}$, we call $\bigcup_{n=0}^
{\infty} X_n$ the {\em result} of the computation $t$ and denote it by
$R_t$.
\ \ $\proofbox$
\end{definition}

Our stipulations that $P$-computations have length $\omega$ does
not restrict the generality. Since atom sets of ac-atoms are finite,
if a clause is applicable with respect to the result of the computation,
it is applicable at some step $n$ of the computation. Consequently, 
like in the case of normal Horn programs, all
possible results of computations of arbitrary transfinite lengths can be 
reached in $\omega$ steps, 
even in the case of infinite programs.

Results of computations are supported models. 

\begin{theorem}
\label{head}
Let $P$ be a Horn $\cal F$-program and let $t$ be a $P$-computation.
Then $R_t$ is a supported model of $P$, that is, $R_t$ is a model of 
$P$ and $R_t \subseteq \hs(P(R_t))$.
\end{theorem}
\begin{proof}
Let $t=(X_n)_{n=0,1,\ldots}$. Clearly $X_0=\emptyset \subseteq
\hs(P(R_t))$. Let $n$ be a non-negative integer. Since $X_{n+1}\in
\tn_P(X_n)$, $X_{n+1}\subseteq \hs(P(X_n))$. Since $P$ is a Horn $\cal
F$-program, it follows that if $r\in P$, $X\subseteq Y$ and $X\models 
\bd(r)$, then $Y\models \bd(r)$. Thus, since $X_n\subseteq R_t$, we have
\[
X_{n+1}\subseteq \hs(P(X_n)) \subseteq \hs(P(R_t)). 
\]
By induction, $R_t=\bigcup_{n=0}^{\infty} X_n \subseteq \hs(P(R_t))$. 

Conversely, let us consider a clause $r\in P$. If $R_t\not\models\bd(r)$ then
$R_t\models r$. Let us then assume that $R_t\models \bd(r)$. Since 
$r$ has finitely many $\cal F$-atoms in the body, and since each $\cal 
F$-atom is of the form $C(X)$, where $X$ is {\em finite}, there is a 
non-negative integer $i$ such that $X_i\models\bd(r)$. By the definition 
of a $P$-computation, $X_{i+1}\in\tn_P(X_i)$. Thus, $X_{i+1}\models \hd
(r)$ and, since $X_{i+1} \subseteq R_t$, $R_t\models \hd(r)$ (by the 
monotonicity of $\hd(r)$). It follows that $R_t\models r$ in the case
when $R_t\models\bd(r)$, as well. Thus, $R_t$ is a model of $P$. 
Since $R_t$ is a model of $P$ and $R_t \subseteq 
\hs(P(R_t))$, $R_t$ is a supported model of $P$.
\end{proof}

We will now show that having a model (being consistent) is a necessary
and sufficient condition for a Horn $\cal F$-program to have a
computation. To this end, we will first introduce a certain class of 
computations.

\begin{definition}
\label{def-new}
Let $M$ be a model of $P$. A {\em canonical $P$-computation with respect
to $M$} is a sequence $t^{P,M}=(X^{P,M}_n)_{n=0,1,\ldots}$ defined as 
follows:
\begin{enumerate}
\item $X^{P,M}_0=\emptyset$ and, 
\item $X^{P,M}_{n+1} = \hs(P(X^{P,M}_n))\cap M$, for every $n\geq 0$.
\ \ $\proofbox$
\end{enumerate}
\end{definition}

We observe that canonical computations involve no nondeterminism. At
each stage there is exactly one way in which we can continue. This
continuation is determined by the model $M$. Before 
we proceed further, we illustrate the concept of a canonical
computation with a simple example.

\begin{example}
Let us assume that $\At=\{a,b,c,d\}$ and let us consider a Horn program 
with cardinality constraints, say $P$, consisting of the following 
clauses:
\begin{quote}
\noindent
$r_1=\ \ 1\{a,d\} \leftarrow 2\{b,d\}$\\
$r_2=\ \ 1\{b,c\} \leftarrow$ \\
$r_3=\ \ 1\{a\} \leftarrow 2\{b,c\}$
\end{quote}
Let $M=\{a,b,c,d\}$. It is easy to check that $M$ is a model of $P$
(it also follows from Proposition \ref{model-exists}, as the constraint 
atoms in the heads of clauses in $P$ are consistent).

We will now construct a canonical $P$-computation with respect to $M$. 
By the
definition $X_0^{P,M}=\emptyset$. The only $X_0^{P,M}$-applicable clause
in $P$ is $r_2$. Since $\{b,c\}\cap M=\{b,c\}$, $X_1^{P,M}=\{b,c\}$.
The clauses $r_2$ and $r_3$ are $X_1^{P,M}$-applicable and $r_1$ is not.
Since $\{a\}\cap M=\{a\}$ and $\{b,c\}\cap M=\{b,c\}$, $X_2^{P,M}=\{a,
b,c\}$. Since $r_2$ and $r_3$ are the only $X_2^{P,M}$-applicable clauses
in $P$, it follows that $X_k^{P,M} = X_2^{P,M}$, for $k=3,4,\ldots$.

By the definition, the union of all sets in the canonical computation 
is included in $M$. Our example demonstrates that canonical 
computations with respect to $M$, in general, do not reconstruct all of 
$M$. \ \ $\proofbox$
\end{example}

The use of the term $P$-computation in Definition \ref{def-new} is 
justified. The following theorem shows that the sequence $t^{P,M}$ is
indeed a $P$-computation. 

\begin{theorem}\label{comp}
Let $P$ be a Horn $\cal F$-program and let $M\subseteq \At$ be a model 
of $P$. Then the sequence $t^{P,M}$ is a $P$-computation.
\end{theorem}
\begin{proof}
We need to show that the conditions (1) and (2) from the definition of 
a $P$-computation hold for the sequence $t^{P,M}$. To prove (1), we 
proceed by induction on $n$. For $n=0$, the condition (1) is, clearly,
satisfied. Let us assume that for some non-negative integer $n$, 
$X^{P,M}_n\subseteq X^{P,M}_{n+1}$ holds. Then
\[
\hs(P(X^{P,M}_n)) \subseteq \hs(P(X^{P,M}_{n+1})).
\]
It follows that
\[
X^{P,M}_{n+1} = \hs(P(X^{P,M}_n))\cap M \subseteq
\hs(P(X^{P,M}_{n+1}))\cap M = X^{P,M}_{n+2}.
\]

To prove (2), let us consider a non-negative integer $n$. By the
definition, $X^{P,M}_{n+1}\subseteq \hs(P(X^{P,M}_n))$. It remains to
prove that $X^{P,M}_{n+1}\models P(X^{P,M}_n)$. Let $r\in P(X^{P,M}_n)$. 
Then $X^{P,M}_n\models \bd(r)$ and, since $X^{P,M}_n\subseteq M$, $M
\models \bd(r)$. We recall that $M$ is a model of $P$. Thus, $M\models 
\hd(r)$. It follows that $M\cap \hs(r) \models \hd(r)$ and, consequently, 
$M\cap \hs(P(X^{P,M}_n))\models \hd(r)$. Since $X^{P,M}_{n+1}= M\cap 
\hs(P(X^{P,M}_n))$, it follows that $X^{P,M}_{n+1} \models P(X^{P,M}_n)$.
\end{proof}

We now have the following corollary to Theorems \ref{head} and \ref{comp}
that characterizes Horn $\cal F$-programs that have computations.

\begin{corollary}
Let $P$ be a Horn $\cal F$-program. Then, $P$ has a model if and only if
it has a $P$-computation. In particular, every Horn $\cal F$-program $P$
without constraint clauses possesses at least one $P$-computation.
\end{corollary}
\begin{proof}
If $M$ is a model of $P$ then the canonical computation $t^{P,M}$
is a $P$-computation (Theorem \ref{comp}). Conversely, if $P$ has a
$P$-computation $t$, then $R_t$ is a model of $P$ (Theorem \ref{head}).
The second part of the assertion follows from the fact that
Horn $\cal F$-programs without constraint clauses have models
(Proposition \ref{model-exists}).
\end{proof}

We use the concept of a computation to identify a certain class of
models of Horn $\cal F$-programs.

\begin{definition}
Let $P$ be a Horn $\cal F$-program. We say that a set of atoms $M$
is a {\em derivable model} of $P$ if there exists a $P$-computation
$t$ such that $M = R_t$.
\ \ $\proofbox$
\end{definition}

Derivable models play in our theory a role analogous to that of the
least model of a normal Horn program. The basic analogy is that
they are the results of bottom-up computations, as is the case for 
the least model of a normal Horn program. 

Theorems \ref{head} and \ref{comp} entail several properties of Horn 
$\cal F$-programs, their computations and models. We gather them in the 
following corollary. Properties (1) and (3) - (6) generalize properties 
of the least model of a normal Horn logic program. 

\begin{corollary} \label{p.sum}
Let $P$ be a Horn $\cal F$-program. Then:
\begin{enumerate}
\item If $P$ is consistent then $P$ has at least one derivable model.
\item For every model $M$ of $P$ there is a largest derivable model 
$M'$ of $P$ such that $M'\subseteq M$.
\item A model $M$ of $P$ is derivable if and only if $M=R_{t^{P,M}}$.
\item If $P$ contains no constraint clauses then $P$ has a largest 
derivable model.
\item Every minimal model of $P$ is derivable.
\item Every derivable model of $P$ is a supported model of $P$.
\end{enumerate}
\end{corollary}
\begin{proof}
(1) Since $P$ has a model, it has a $P$-computation (Theorem
\ref{comp}). The result of this computation is a model of $P$
(Theorem \ref{head}). By the definition, this model is
derivable.

\noindent
(2) Let $M$ be a model of $P$ and let $t=(X_n)_{n=0,1,\ldots}$ be the
canonical $P$-computation for $M$. Then, $R_t$ is a derivable model of
$P$ and $R_t\subseteq M$. Let $s =(Y_n)_{n=0,1,\ldots}$ be a
$P$-computation such that $R_s\subseteq M$. Clearly, we have
$Y_0\subseteq
X_0$. Let us consider an integer $n>0$ and let us assume that 
the inclusion $Y_{n-1} \subseteq X_{n-1}$ holds. 
Since $R_s\subseteq M$, $Y_n\subseteq M$. Thus, by
the definition of a $P$-computation,
\[
Y_n\subseteq \hs(P(Y_{n-1}))\cap M.
\]
Since $P$ is a Horn $\cal F$-program and since we have
$Y_{n-1}\subseteq
X_{n-1}$, $\hs(P(Y_{n-1})) \subseteq \hs(P(X_{n-1}))$. Thus,
\[
Y_n\subseteq \hs(P(X_{n-1}))\cap M= X_n.
\]
It follows now by induction that $R_s\subseteq R_t$. Thus, $R_t$ is the
largest derivable model contained in $M$.

\noindent
(3) Let $M$ be a model of $P$. The argument we used in (2) shows that
the result of the canonical computation from $P$ with respect to $M$ is
the greatest derivable model contained in $M$. If $M$ is derivable,
then $M = R_{t^{P,M}}$. The converse implication follows by the
definition.

\noindent
(4) The set $\At$ is a model of $P$. Let $R$ be the result of the
canonical $P$-computation for $\At$. Clearly, $R$ is a derivable
model of $P$. We will show that every derivable model of $P$ is a subset
of $R$. Let $M$ be a derivable model of $P$. By (3), $M$ is the result of
a canonical computation for $M$. Since $M\subseteq \At$, it follows
by an induction argument that for every non-negative integer $n$, $
X^{P,M}_n \subseteq X^{P,\At}_n$ (we omit the details, as the argument
is similar to that in the proof of (2)). Consequently, $M\subseteq R$.

\noindent
(5) This assertion follows directly from (2). 

\noindent
(6) This assertion follows directly from Theorem \ref{head}.
\end{proof}

Despite analogies with the least model of a normal Horn program,
derivable models are not, in general, minimal. For instance, a
program with cardinality constraints consisting of a single clause
\[
1\{a,b\}\leftarrow
\]
has three derivable models: $\{a\}$, $\{b\}$ and $\{a,b\}$, only two of
which are minimal.

Horn $\cal F$-programs generalize Horn normal logic programs (see
Section \ref{nlp-mca} for details.). For a Horn $\cal F$-programs 
without constraint clauses, the canonical computation with respect to 
the set of all atoms is a counterpart to the bottom-up computation 
determined by a normal Horn program.

\section{Stable models of monotone $\cal F$-programs}\label{stable}

We will now use the results of the two previous sections to introduce and
study the class of {\em stable} models of monotone $\cal F$-programs.

\begin{definition}\label{d-stable}
Let $P$ be a monotone $\cal F$-program and let $M\subseteq \At$. The 
{\em 
reduct} of $P$ with respect to $M$, $P^M$ in symbols, is a Horn $\cal 
F$-program obtained from $P$ by (1) removing from $P$ every $\cal 
F$-clause containing in the body a literal $\n(A)$ such that $M\models 
A$, and (2) removing all literals of the form $\n(A)$ from all the 
remaining clauses in $P$. A set of atoms $M$ is a {\em stable} model of 
$P$ if $M$ is a derivable model of the reduct $P^M$.
\ \ $\proofbox$
\end{definition}

The following result is easy to show (and so we omit its proof) 
but useful.

\begin{lemma}\label{l.reduct}
Let $P$ be a monotone $\cal F$-program. If $M$ is a model of $P$, then
$M$ is a model of $P^M$. \ \ $\proofbox$
\end{lemma}

\begin{example}
\label{ex-stab}
We illustrate the concept of stable models of monotone 
$\cal F$-programs with
examples underlining some aspects of their properties. The class $\cal 
F$ we use in this example consists of {\em all} cardinality
constraints which, we recall, are monotone (Section
\ref{basic-concepts}).

Let $P$ be a program consisting of the following two clauses:
\begin{quote}
\noindent
$1\{a,b\} \lar 1\{d\}, \n (1\{b,c\})$\\
$1\{a,d\} \lar$
\end{quote}
We will now investigate properties of some sets with respect to this
program.\\
(1) The set $M_1 = \emptyset$ is not a model of our program $P$. As we
will see soon (Proposition \ref{m-stable}), stable models are supported 
models and, consequently,
also models. Thus $\emptyset$ is not a stable model of $P$.\\
(2) The set $M_2 = \{a,b,c\}$ is a model of $P$. But $M_2$ is not
a stable model of $P$. Indeed, let us compute $P^{M_2}$. It consists of
just one clause: $1\{a,d\} \lar$. Since $M_2$ is not a derivable model
of $P^{M_2}$ (it contains an atom not occurring in any head of the
clause of $P^{M_2}$), $M_2$ is not a stable model of $P$\\
(3) The set $M_3 = \{a,d\}$ is a stable model of $P$. The reduct
$P^{M_3}$ consists of two clauses:
\begin{quote}
\noindent
$1\{a,b\} \lar 1\{d\}$\\
$1\{a,d\} \lar$
\end{quote}
The sequence $\emptyset,\{a,d\},\{a,d\},\ldots$ is a $P^{M_3}$-computation.
Thus, $M_3$ is a derivable model of $P^{M_3}$ and hence $M_3$ is a
stable model of $P$\\
(4) The set $M_4 = \{a\}$ is a stable model of $P$. The reduct
$P^{M_4}$ consists of two clauses:
\begin{quote}
\noindent
$1\{a,b\} \lar 1\{d\}$\\
$1\{a,d\} \lar$
\end{quote}
The sequence $\emptyset,\{a\},\{a\},\ldots$ is a
$P^{M_4}$-computation. Thus $\{a\}$ is a stable model of $P$.

In our example $M_4 \subset M_3$. Thus, in contrast to normal logic
programs (but not to lparse programs), stable models of abstract 
constraint programs can nest. That is, they do not satisfy the
antichain (minimality with respect to inclusion) property.

The program $P$ that we considered above has stable models. It is easy to
construct examples of programs that have no stable models. For
instance, a program consisting of just one clause: $2\{a,b,c\} \lar 
\n(1\{a,b\})$ has models but no stable models.  \ \ $\proofbox$
\end{example}

Stable models of a monotone $\cal F$-program $P$ are indeed models of $P$. 
Thus, the use of the term ``model'' in their name is justified. In fact, 
a stronger property holds: stable models of monotone $\cal F$-programs are 
supported. This again generalizes a well-known property of normal logic 
programs\footnote{Incidentally, in the case of programs with weight 
constraints in the lparse syntax, no such property has been 
established as supported models have not been defined for that 
formalism.}.

\begin{proposition}\label{m-stable}
Let $P$ be a monotone $\cal F$-program. If $M\subseteq \At$ is 
a stable model of 
$P$ then $M$ is a supported model of $P$.
\end{proposition}
\begin{proof}
First, let us observe that it follows directly from the corresponding 
definitions
that $\tn_P(M)=\tn_{P^M}(M)$. Next, since 
the set $M$ is a derivable model of
$P^M$, $M$ is a supported model of $P^M$ (Corollary \ref{p.sum}(6)).
Thus, by Theorem \ref{ch-sup}, $M\in \tn_{P^M}(M)$ and, consequently, 
$M\in \tn_{P}(M)$. It follows that $M$ is a supported model of $P$.
\end{proof}

With the notion of a stable model in hand, we can strengthen
Theorem \ref{head}.

\begin{theorem} \label{der-stb}
Let $P$ be a Horn $\cal F$-program. A set of atoms $M\subseteq \At$ is 
a derivable model of $P$ if and only if $M$ is a stable model of $P$.
\end{theorem}
\begin{proof}
The assertion is a direct consequence of the fact that for
every Horn $\cal F$-program $P$ and for every set of atoms $M$, 
$P=P^M$.
\end{proof}

We will now prove yet another result that generalizes a property of 
stable models of normal logic programs (cf. work on extending the
semantics of stable models to logic programs with integrity 
constraints \cite{li96}).  

\begin{proposition}\label{prop.comp}
Let $P$ and $Q$ be two monotone $\cal F$-programs. 
\begin{enumerate}
\item If $M$ is a stable model of $P$ and a model of $Q$ then $M$ is a 
stable model of $P \cup Q$.
\item If $Q$ consists of constraint clauses and $M$ is a stable model 
of $P\cup Q$ then $M$ is a stable model of $P$.
\end{enumerate}
\end{proposition}
\begin{proof}
(1) Since $M$ is a stable model of $P$, $M$ is a derivable model of $P^M$.
By Corollary \ref{p.sum}(3), $M$ is the result of the canonical 
$P^M$-computation with respect to $M$. Since $M$ is a model of $P\cup Q$, 
by Lemma \ref{l.reduct} $M$ is a model of $(P\cup Q)^M
=P^M\cup Q^M$. Therefore, the canonical 
$(P^M\cup Q^M)$-computation with respect to $M$ is well defined. Its
result is clearly contained in $M$. On the other hand, it contains the 
result of the canonical $P^M$-computation with respect to $M$, which is 
$M$. Therefore, the result of the canonical $(P^M\cup Q^M)$-computation
with respect to $M$ is $M$. Thus, $M$ is a derivable model of $(P\cup 
Q)^M$ and a stable model of $P\cup Q$.

\noindent
(2) Since $M$ is a stable model of $P\cup Q$, $M$ is the result of
a $(P\cup Q)^M$-computation, say $t$. Since $Q$ consists of constraint 
clauses, $t$ is a $P^M$-computation (constraint clauses, having
inconsistent heads, do not participate in computations). Thus,
$M$ is also a result of a $P^M$-computation, that is, $M$ is a stable 
model of $P$.
\end{proof}

\section{Monotone $\cal F$-programs and normal logic 
programming}\label{nlp-mca}

The main goal of this paper is to propose a way to integrate abstract 
constraints into normal logic programming. In this section, we show that 
our formalism of $\cal F$-programs {\em contains} 
normal logic programming (modulo a very simple encoding) so that all 
major two-valued semantics are preserved.

To this end, let us consider an abstract constraint:
\[
PB = \{X\subseteq \At \colon X\not=\emptyset \}.
\]
We note that $PB$ is identical with the
pseudo-boolean constraint (we introduced 
pseudo-boolean constraints in Section \ref{basic-concepts}):
\[
PB = PB(1,W,\leq),
\]
where $W$ is a weight function on $\At$ such that $W(a)=1$, for every
$a\in \At$. Clearly, the constraint $PB$ is monotone.
We will show that normal logic programs can be encoded
as $\{PB\}$-programs or, more generally, as monotone 
$\mathcal{F}$-programs,
for every class $\mathcal{F}$ of monotone abstract constraints such
that $PB\in \mathcal{F}$. In what follows, if $a\in \At$, we 
will write $PB(a)$ for a $\{PB\}$-atom $PB(\{a\})$. 

We note that for every $a\in \At$ and every interpretation $M\subseteq 
\At$, $M\models a$ if and only if $M\models PB(a)$. That is, a 
propositional atom $a$ is logically equivalent to an abstract constraint 
atom $PB(a)$. This equivalence suggests an encoding of a normal logic 
program $P$ as $\{PB\}$-program $P^{pb}$. Namely, if r is a normal logic
program clause
\[
a \leftarrow b_1,\ldots,b_m,\n(c_1),\ldots,\n(c_n)
\]
we define $r^{pb}$ to be a $\{PB\}$-clause
\[
PB(a) \leftarrow PB(b_1),\ldots,PB(b_m),\n(PB(c_1)),\ldots,\n(PB(c_n)).
\]
For a normal logic program $P$, we define $P^{pb}=\{r^{pb}\colon r\in 
P\}$. By our earlier comments, $P^{pb}$ is a {\em monotone} $\cal 
F$-program, for every class of monotone constraint atoms containing
the constraint $PB$. 

We note that due to the equivalence of $a$ and $PB(a)$, which we discussed
above, for every interpretation $M\subseteq \At$ we have 
\begin{equation}
\label{eq23c}
M\models \bd(r)\ \ \mbox{if and only if}\ \ M\models \bd(r^{pb}).
\end{equation}
(here and in other places we use symbols such as $\bd(r)$, $\hd(r)$ and 
$\hd(P)$ also in the context of normal logic programs, and assume their 
standard meaning). 

Our first result involves operators associated with programs. 
Let $P$ be a normal
logic program. We recall that the one-step provability operator $\tp_P$ 
\cite{vEK76} is defined as follows: for every $M\subseteq \At$,
\[
\tp_P(M)=\{\hd(r)\colon r\in P\ \mbox{and}\ M\models\bd(r)\}.
\]
We have the following basic property of the translation $P\mapsto P^{pb}$.

\begin{proposition}
\label{tptn}
Let $P$ be a normal logic program. Then for every $M\subseteq \At(P)$,
$\tn_{P^{pb}}(M) = \{\tp_P(M)\}$.
\end{proposition}
\begin{proof}
We will write $r$ and $r'$ for a pair of corresponding clauses in $P$ and
$P^{pb}$. That is, if $r\in P$ then $r'=r^{pb}$ is the counterpart of
$r$ in $P^{pb}$. Conversely, if $r'\in P^{pb}$, $r$ is the clause in $P$ 
such that $r^{pb}=r'$. Clearly, we have $\hs(r') =\{\hd(r)\}$.

By the equivalence (\ref{eq23c}), a clause $r\in P$ is $M$-applicable if
and only if $r'$ is $M$-applicable. Thus, we have 
\begin{equation}
\label{eq23z}
\hs(P^{pb}(M)) = \hd(P(M)) = \tp_P(M).
\end{equation}
Let $r'\in P^{pb}(M)$ and let $a=\hd(r)$. It follows that $r\in P(M)$
and $a\in \tp_P(M)$. Since $\hd(r')=PB(a)$, $\tp_P(M)\models \hd(r')$.
Thus, $\tp_P(M)$ is one-step nondeterministically provable from $M$ and
$P^{pb}$, that is, $\tp_P(M)\in\tn_{P^{pb}}(M)$.

Next, let us consider $M'\in \tn_{P^{pb}}(M)$. By the definition,
$M'\subseteq \hs(P^{pb}(M))$. Thus, by (\ref{eq23z}), we have $M'
\subseteq \tp_P(M)$. Let us now consider $a\in \tp_P(M)$. It follows 
that there is a clause $r\in P(M)$ such that $\hd(r)=a$. Consequently, 
$r'\in P^{pb}(M)$ and $\hd(r')=PB(a)$. Since $M'\in \tn_{P^{pb}}(M)$,
$M'\models \hd(r')$. Thus, $a\in M'$. It follows that
$M'=\tp_P(M)$ and, consequently, $\tn_{P^{pb}}(M) = \{\tp_P(M)\}$.
\end{proof}

This result entails a proposition concerning Horn programs.

\begin{proposition}
\label{v.new}
Let $P$ be a normal Horn logic program. Then $M$ is a least model of $P$
if and only if $M$ is a derivable model of $P^{pb}$.
\end{proposition}
\begin{proof}
We first observe that the sequence $\{\tp_P\uparrow n
(\emptyset )\}_{n=0,1,\ldots}$ is
a $P^{pb}$-computation (one can show this by an easy inductive argument, 
using the relationship between $\tp_P$ and $\tn_{P^{pb}}$ established 
by Proposition \ref{tptn}). Since $M$ is the limit of the sequence 
$\{\tp_P\uparrow n(\emptyset)\}_{n=0,1,\ldots}$, $M$ is a derivable model of 
$P^{pb}$.

Conversely, if $M$ is a derivable model of $P^{pb}$, then $M$ is the
result of a derivation $\{X_n\}_{n=0,1,\ldots}$ from $P^{pb}$. Thus,
for every $n=0,1,\ldots$, $X_{n+1}\in \tn_{P^{pb}}(X_n)$. By Proposition
\ref{tptn}, $X_{n+1}=\tp_P(X_n)$. Since $X_0=\emptyset$, it follows 
that for every $n=0,1,\ldots$, $X_n=\tp_P\uparrow n(\emptyset)$. Consequently,
$M=\bigcup_{n=0}^\infty \tp_P\uparrow n(\emptyset)$ and so, $M$ is the least
model of $P$.
\end{proof}

We can now prove the main result of this section demonstrating that
the embedding $P\mapsto P^{pb}$ preserves all the semantics considered
in the paper.

\begin{theorem}
\label{lp-mca}
Let $P$ be a normal logic program and let $M$ be a set of atoms.
Then $M$ is a model (supported model, stable model) of $P$ if and
only if $M$ is a model (supported model, stable model) of $P^{pb}$.
\end{theorem}
\begin{proof}
It is well known that $M$ is a model of $P$ if and only if
$\tp_P(M)\subseteq M$ \cite{ap90}. By Proposition \ref{tptn}, the latter 
condition is equivalent to the condition that there is $M'\in 
\tn_{P^{pb}}(M)$ such that $M'\subseteq M$. By Theorem \ref{model}, 
this last condition is equivalent to $M$ being a model of $P^{pb}$.
Thus, $M$ is a model of $P$ if and only if $M$ is a model of $P^{pb}$.

The proof for the case of supported models is essentially the same. It
relies on the fact that $M$ is a supported model of $P$ if and only
if $M=\tp_P(M)$ \cite{ap90} and uses Proposition \ref{tptn} and Theorem
\ref{ch-sup}.

Let us assume now that $M$ is a stable model of $P$. It follows that
$M$ is the least model of $P^M$. By Proposition \ref{v.new}, $M$ is a
derivable model of $[P^M]^{pb}$. It follows from the definitions of the
reducts of normal logic programs and $\{PB\}$-programs that $[P^M]^{pb}=
[P^{pb}]^M$. Thus, $M$ is a stable model of $P^{pb}$. The converse
implication can be proved in the same way.
\end{proof}

There are other ways to establish a connection between normal logic 
programs and programs with abstract constraints. We will now define a 
class of monotone $\cal F$-programs, which offers a most direct 
extension of normal logic programming.

\begin{definition}
\label{ddet}
An $\cal F$-atom $C(X)$ is {\em definite} if $X$ is a minimal
element in $C$. An $\cal F$-clause $r$ is {\em definite} if
$\hd(r)$ is a definite $\cal F$-atom. An $\cal F$-program is {\em
definite} if every clause in $P$ is definite.
\ \ $\proofbox$
\end{definition}

We use the term definite following the logic programming tradition
(cf. \cite{vEK76}, for instance), where it is used for clauses
whose heads provide ``definite'' information (as opposed to being
disjunctions and so listing several possible alternatives).

\begin{example}
Let $\cal F$ consist of two monotone constraints, $C_1$ and $C_2$ where:
\[
C_1 = \{X\subseteq \At \colon \mbox{$\{a,b\} \subseteq X$ or $\{a,c\} \subseteq X$
or $|X|$ is infinite}\}
\]
and 
\[
C_2 = \{X \subseteq \At\colon \mbox{$\{d,e\} \subseteq X$} \}.
\]
The constraint $C_1$ has two minimal elements: $\{a,b\}$ and $\{a,c\}$. 
The constraint $C_2$ has just one minimal element: $\{d,e\}$.

These two monotone constraints generate the following
three definite atoms: $C_1(\{a,b\})$, 
$C_1(\{a,c\})$, and $C_2(\{d,e\})$. An $\cal F$-program consisting of
the following clauses is definite:
\begin{quote}
\noindent
$C_1(\{a,b\}) \leftarrow$\\
$C_1(\{a,c\}) \leftarrow C_1(\{a,b,c\}),\n(C_2(\{a,b,d,e\}))$\\
$C_2(\{d,e\}) \leftarrow$
\end{quote}

We note that some monotone constraints do not yield any definite 
constraint atoms. It happens when they have no finite minimal elements.
A constraint $C$ consisting of all infinite subsets of $\At$ offers a
specific example.
\ \ $\proofbox$
\end{example}

Definite $\cal F$-atoms have the following simple properties.

\begin{proposition}
\label{det}
Let $X\subseteq \At$ and let $C(X)$ be a definite $\cal F$-atom. 
Then $C(X)$ is consistent and, for every $M\subseteq \At$, $M\models 
C(X)$ if and only if $X \subseteq M$.
\end{proposition}
\begin{proof}
If $M\models C(X)$ then $M\cap X\in C$. Since $C(X)$ is a definite
$\cal F$-atom, $X$ is a minimal element in $C$. It follows that $M\cap X
= X$ and so, $X\subseteq M$. Conversely, if $X\subseteq M$ then $M\cap
X = X$. Since $X\in C$, $M\cap X\in C$. Thus, $M\models C(X)$. This
argument proves the second part of the assertion. In particular, it
follows that $X\models C(X)$. Thus, $C(X)$ is consistent.
\end{proof}

The intuition behind the notion of a definite $\cal F$-atom is now
clear. Given a definite $\cal F$-program and an interpretation $M$,
there is always a way to satisfy the heads of all $M$-applicable clauses
(due to consistency of definite $\cal F$-atoms). Moreover, there is 
only {\em one} way to do so if we want only to use atoms appearing in 
the headsets of $M$-applicable clauses (due the the second property from
Proposition \ref{det}). Thus, computing with definite $\cal
F$-programs does not involve nondeterminism. Indeed, we have the 
following result.

\begin{proposition}
\label{det1}
Let $P$ be a definite $\cal F$-program. Then, for every set of
atoms $M$, $|\tn_P(M)|=1$.
\end{proposition}
\begin{proof}
Let $r\in P(M)$. Since $\hd(r)$ is a definite $\cal F$-atom, then 
$\hs(r)\models \hd(r)$. We now observe that $\hs(r)\subseteq\hs(P(M))$.
Thus, for every $r\in P(M)$, $\hs(P(M))\models \hd(r)$. By the
definition of the one-step nondeterministic provability, $\hs(P(M))\in 
\tn_P(M)$. Thus, $|\tn_P(M)|\geq 1$.

Next, let us consider $M'\in \tn_P(M)$. From the definition of $\tn_P(M)$,
it follows that $M'\subseteq \hs(P(M))$. To prove the converse inclusion,
let $r\in P(M)$. Again by the definition of $\tn_P(M)$, we have that $M'
\models \hd(r)$. Since $\hd(r)$ is a definite $\cal F$-atom, 
Proposition \ref{det} implies that $\hs(r)\subseteq M'$. Thus, $\hs(P(M))
\subseteq M'$.

It follows that $\hs(P(M))=M'$ and so, $|\tn_P(M)| = 1$. 
\end{proof}

Thus, for a definite $\cal F$-program $P$, the operator $\tn_P$ is
deterministic and, so, can be regarded as an operator with both the
domain and codomain ${\cal P}(\At)$. We will write $\td_P$, to denote
it. Models, supported models and stable models of a definite monotone
$\cal F$-program (for supported models we do not need the monotonicity
assumption) can be introduced in terms of the operator $\td_P$ in {\em 
exactly} the same way the corresponding concepts are defined in normal 
logic programming. In particular, the algebraic treatment of 
logic programming developed in \cite{fi99,przy90,dmt00a} applies
to definite $\cal F$-programs and results in a natural
and direct extension of normal logic programming. We note that this 
comment extends to 3- and 4-valued semantics of partial models,
supported models and stable models (including the Kripke-Kleene
semantics and the well-founded semantics)\footnote{Results in
\cite{dpb01,pdb04,p04} are related to this observation. They
concern programs with aggregates, whose clauses have heads consisting of
{\em single} atoms and so, are definite.}.

We will explicitly mention just one result on definite monotone $\cal
F$-programs (in fact, definite Horn programs) here, as it will be used 
later in the paper.

\begin{proposition}
\label{horn-det}
Let $P$ be a definite Horn $\cal F$-program. Then $P$ has exactly 
one derivable model and this model is the least model of $P$.
\end{proposition}
\begin{proof}
Since $P$ is definite, it contains no constraint clauses and so, 
it has a model (Proposition \ref{model-exists}).
Thus, it has at least one $P$-computation. Let $(X_n)_{n=0,1,\ldots}$ 
and $(Y_n)_{n=0,1,\ldots}$ be two $P$-computations. By the definition,
$X_0=\emptyset=Y_0$. Let us assume that for some $n\geq 0$, $X_n=Y_n$.
By the definition of $P$-computations, 
\[
X_{n+1}\in \tn_P(X_n)\ \ \mbox{and}\ \ Y_{n+1}\in \tn_P(Y_n).
\]
By the induction hypothesis, $X_n=Y_n$. Thus, $\tn_P(X_n)=\tn_P(Y_n)$.  
Since $P$ is definite, $|\tn_P(X_n)|=|\tn_P(Y_n)|=1$ and so, 
$X_{n+1}=Y_{n+1}$. Thus,
both computations coincide and $P$ has exactly one $P$-computation and 
so, exactly one derivable model. Since every model of $P$ contains a
derivable model, it follows that the unique derivable model of $P$ is
also a least model of $P$.
\end{proof}

\section{Encoding lparse programs as monotone
$\cal F$-programs}
\label{mca-cc}

We will now investigate the relation between lparse programs
\cite{nss99,sns02} and programs with monotone abstract constraints.
We start by reviewing the syntax and the semantics of lparse
programs.

A {\em weighted set of literals} is a function $W:X\rightarrow\{\ldots,
-1,0,1,\ldots \}$, where $X\subseteq \At \cup \{\n(a) \colon a \in \At\}$ 
is finite. We call $X$ the {\em literal set} of $W$ and denote it by 
$\lst(W)$. The set of atoms that appear in literals in $\lst(W)$ is the 
{\em atom set} of $W$. We denote it by $\as(W)$. Sometimes it will be 
convenient to write $W$ explicitly. To this end, we will write $W$ as
\begin{equation}
\{a_1 = w_1, \ldots, a_m = w_m, \n(b_1)=w'_1,\ldots,\n(b_n)=w'_n\},
\label{eq:latom}
\end{equation}
where the domain of the function $W$ is $\{a_1,\ldots,
a_m,\n(b_1),\ldots, \n(b_n) \}$, and $w_1
= W(a_1),\ldots, w'_n = W(\n(b_n))$, loosely following the lparse
notation.  Thus, when the domain of $W$ is $\{a,b,c\}$ and $W(a) = 1, 
W(b) = 2$, and $W(c) = 1$, then we write $W$ as $\{a =1,b=2,c=1\}$. 

An {\em lparse atom} ({\em l-atom}, for short) is an expression of
the form $kWl$, where $W$ is a weighted set of literals, and $k$ and $l$ 
are integers such that $k \leq l$. By the {\em literal set} of 
an l-atom $A=kWl$ we mean $\lst(W)$ and write $\lst(A)$ to denote 
it (in a similar way, we extend the definition and the notation of the
atoms set to the case of l-atoms).

We say that a set of atoms (interpretation) $M$ satisfies an l-atom 
$kWl$ if 
\[
k\leq\sum_{\scriptsize{\begin{array}{c}p\in \lst(W)\\ p\in M\end{array}}} 
W(p) + \sum_{\scriptsize{\begin{array}{c}\n(p)\in \lst(W)\\ p\not\in M
\end{array}}} W(p) \leq l
\]
($M\models kWl$, in symbols). 
We note that it is easy to give an example of an inconsistent l-atom.
For instance, $2\{a=1\}2$ is inconsistent. We will use $I$ to denote
any inconsistent constraint (it does not matter which, as all are
equivalent to each other).

An {\em lparse clause} (l-clause, for short) is an expression
$r$ of the form
\[
 A \leftarrow B_1,\ldots, B_n,
\]
where $A$ and $B_i$, $1\leq i\leq n$, are l-atoms. We call $A$ the 
{\em head}
of $r$ and $\{B_1,\ldots,B_n\}$ the {\em body} of $r$. We denote them
by $\hd(r)$ and $\bd(r)$, respectively. An {\em lparse program} is a 
finite set of l-clauses.

We say that a set $M\subseteq \At$ {\em satisfies} an l-clause $r$ if
$M$ satisfies $\hd(r)$ whenever it satisfies each l-atom in the body of
$r$. We say that $M$ satisfies an lparse program $P$ if $M$ satisfies 
each l-clause in $P$. We write $M\models r$ and $M\models P$ in these
cases, respectively.

We note that lparse programs allow both negative literals and negative 
weights in l-atoms. However, in \cite{sns02} it is argued that negative 
weights can be expressed using negative literals and vice versa and, 
hence, one is inessential when the other is available. In fact, 
in~\cite{sns02} an l-atom with negative weights is treated simply as a 
shorthand for the corresponding constraint with non-negative weights. 
We follow this approach here and from now on consider only
l-atoms $kWl$, where $W$ assigns non-negative weights to literals. 

Before we continue, let us define $\mathcal{PB}$ to be a set of all 
pseudo-boolean constraints of the form $PB(k,W,\leq)$, where $k$ is a 
non-negative integer and $W$ a weight function assigning to elements 
of $\At$ non-negative integers (cf. Section \ref{basic-concepts}). 
Directly from the definition it follows that every constraint in ${\cal 
PB}$ is monotone. 

Let us consider an l-atom $lW$ which contains no negative literals (and,
as it is evident from the notation, no upper bound). In particular,
$\lst(W)=\as(W)$. Let $W'$ be an extension of $W$, which assigns $0$ to 
every atom $p\in \At \setminus \as(W)$. We observe that a set $M\subseteq
\At$ is a model of $lW$ if and only if $M$ is a model of the 
${\cal PB}$-atom $A(X)$, where $A=PB(l,W',\leq)$ and $X=\as(W)$. 
Therefore, we will regard such an l-atom $lW$ as a ${\cal PB}$-atom or, 
speaking more formally (but with some abuse of notation) we will denote 
by $lW$ the ${\cal PB}$-atom $A(X)$. 

If $W=\{a=1\}$ and $l=2$, then
the corresponding ${\cal PB}$-atom is inconsistent (it is one of many 
inconsistent $\mathcal{PB}$-atoms). As in the case of l-atoms, we will
write $I$ to denote (any) inconsistent $\mathcal{PB}$-atom.

This discussion suggests that lparse programs built of l-atoms without 
negative literals and upper bounds can be viewed as Horn ${\cal
PB}$-programs. We will exploit  
that relationship below in the definition of the reduct, and will 
subsequently extend it to all lparse programs.

Let $P$ be an lparse program and let $M\subseteq \At$. An 
{\em lparse-reduct} of $P$ with respect to $M$ is a 
${\cal PB}$-program obtained by:
\begin{enumerate}
\item eliminating from $P$ every clause $r$ such that $M\not\models B$,
for at least one l-atom $B\in \bd(r)$.
\item replacing each remaining l-clause $r= kWl \lar k_1W_1l_1,\ldots,
k_nW_nl_n$ with ${\cal PB}$-clauses of the form
\[
1\{a=1\} \lar k_1' W_1',\ldots,k_n' W_n',
\]
where $a \in \lst(W) \cap M$, $W_i'$ is $W_i$ restricted to $\lst(W_i)
\cap \At$, and
\[
k_i' = k_i - \sum_{\scriptsize{\begin{array}{c} \n(p) \in \lst(W_i)\\ p \not\in
    M\end{array}}} W_i(\n(p)) 
\]
\end{enumerate}
(by our comments above, expressions of the form $l'W'$ denote
${\cal PB}$-atoms). With some abuse of notation, we denote the reduct of 
$P$ with respect to $M$ by $P^M$ (the type of the program, an lparse 
program or a ${\cal PB}$-program, determines which reduct we have in 
mind). By our comments above, $P^M$ can be regarded as a 
definite Horn 
${\cal PB}$-program. Thus, $P^M$ has a least model, $\lm(P^M)$ 
(Proposition \ref{horn-det}). This model is the result of the canonical 
computation from $P^M$ with respect to $M$.  

\begin{definition}
\label{def-nss}
Let $P$ be an lparse program. A set $M\subseteq \At$ is an 
{\em lparse-stable
model} of $P$ if  $M=\lm(P^M)$ and $M\models P$. \ \ $\proofbox$
\end{definition}

We will now show that {\em all} lparse programs can be viewed as ${\cal
  PB}$-programs. 
This task involves
 two steps. First, we show how to translate lparse
programs to \lparseNegfree{} programs so that lparse-stable models are
preserved. Second, we show that for \lparseNegfree{} programs the two
definitions of stable models presented in the paper
(Definitions \ref{d-stable} and \ref{def-nss}) are equivalent.

An lparse program $P$ can be translated to a
\lparseNegfree{} program $P'$, 
as follows. We recall that by our earlier comments, we need to
consider only lparse programs with no negative weights.
For each negated literal $\n(b)$ appearing in $P$,
introduce a new propositional atom $\bar{b}$ and an l-clause $\bar{b}
\leftarrow 0\{b=1\}0$. 
Then we replace each l-atom $kWl$ 
where the weighted set of literals $W$ is of the form~(\ref{eq:latom})
with an l-atom
\[
k\{a_1=w_1,\ldots,a_m=w_m,\bar{b_1}=w'_1,\ldots,\bar{b_n}=w'_n\}l .
\]

It is straightforward to show that this transformation preserves
lparse-stable models in the following sense. 
\begin{proposition}
Let $P$ be an lparse program, $P'$ a \lparseNegfree{}
program obtained by the translation above, and 
$B$ the set of new atoms introduced in the translation. 
Then,
\begin{itemize}
\item if $M$ is an lparse-stable model of $P$ then $M \cup \{\bar{b}
  \colon b \in B \setminus M\}$ is a lparse-stable model of $P'$ and 
\item if $M'$ is a lparse-stable model of $P'$ then $M = M' \setminus B
  $ is an lparse-stable model of $P$. \ \ $\proofbox$
\end{itemize}
\end{proposition}

Now we show that \lparseNegfree{} programs can be translated to 
${\cal PB}$-programs. 
To simplify the description of the encoding and make it uniform, we
assume that all bounds are present.
Let $r$ be an l-clause
\[
k W l \lar k_1 W_1 l_1,\ldots, k_m W_m l_m.
\]
We represent this l-clause by a {\em pair} of ${\cal PB}$-clauses, 
$e_1(r)$ and $e_2(r)$ that we define as 
\[
k W  \lar k_1 W_1, \ldots, k_m W_m, \n((l_1+1)W_1), \ldots, \n((l_m+1)W_m),
\]
and
\[
I \lar (l+1)W, k_1 W_1, \ldots, k_m W_m, \n((l_1+1)W_1),
\ldots, \n((l_m+1)W_m),
\]
respectively. We recall that the symbol $I$, appearing in the clause 
$e_2(r)$, stands for the inconsistent $\mathcal{PB}$-atom introduced
above. 

Now, given a \lparseNegfree{} program $P$, we translate it into a ${\cal 
PB}$-program
\[
e(P)= \bigcup_{r \in P} \{e_1(r), e_2(r)\}.
\]

\begin{theorem}
\label{ca-mca}
Let $P$ be a \lparseNegfree{} program. A set $M$ is an
lparse-stable 
model of $P$ if and only if $M$ is a stable model of $e(P)$, as defined 
for ${\cal PB}$-programs.
\end{theorem}
\begin{proof}
In the proof we will use the notation:
\[
P_1=\bigcup\{e_1(r)\colon r\in P\}\ \ \mbox{and}\ \ 
P_2=\bigcup\{e_2(r)\colon r\in P\}.
\] 

Let us assume first  that $M$ is an lparse-stable model of a
\lparseNegfree{}
program $P$. We will show that $M$ is a stable model of the 
${\cal PB}$-program $e(P)$, which in our terminology is equal to $P_1\cup P_2$.

Since $M$ is an lparse-stable model of $P$, it is a model of $P$ 
(Definition~\ref{def-nss}). Consequently, $M$ is a model of $P_2$. 
By Proposition \ref{prop.comp} to complete this part of the proof, it 
suffices to show that $M$ is a stable model of the program 
$P_1$. To this end, we note that the definitions of the respective 
reducts imply that a clause
\[
1\{a=1\}\lar k_1W_1,\ldots k_mW_m
\]
belongs to the lparse-reduct $P^M$ if and only if the reduct
$P_1^M$ contains a clause
\[
kW\lar k_1W_1,\ldots k_mW_m
\]
such that $a\in \as(W)$ and $M \models k_i W_i$ for all $1 \leq i \leq
m$.

From this relationship it follows that the results of the canonical 
computations from $P^M$ and $P_1^M$ with respect to $M$ coincide (we 
recall that both reducts are Horn ${\cal PB}$-programs). Since $M$ is 
the least model of $P^M$, it is the result of the canonical 
computation from $P^M$ with respect to $M$. Thus, $M$ is also the result 
of the canonical computation from $P_1^M$ with respect to $M$. In other
words, $M$ is a derivable model of $P_1$ and, consequently, a 
stable model of $P_1$.

Conversely, let us assume that $M$ is a stable model of 
$P_1 \cup P_2$. It follows that $M$ is a model of $P_1\cup P_2$ and, 
consequently, a model of $P$. Next, we note that since $M$ is a 
stable model of $P_1\cup P_2$, it is a stable model of 
$P_1$ (by Proposition \ref{prop.comp}). Thus, it is a derivable model 
of its reduct $P_1^M$ and, therefore, it is also the result of the 
canonical computation from $P_1^M$ with respect to $M$. Our observation
about the relationship between the reducts $P_1^M$ of and $P^M$ holds 
now, as well. Consequently, $M$ is the result of the canonical 
computation from $P^M$ with respect to $M$. Thus, $M$ is a derivable 
model of $P^M$. Since $P^M$ is a definite Horn ${\cal PB}$-program, 
it has only one derivable model --- its least model. It follows that 
$M$ is the least model of $P^M$ and, consequently, an lparse-stable 
model of $P$.
\end{proof}

Theorem \ref{ca-mca} shows that ${\cal PB}$-programs can express arbitrary
\lparseNegfree{} programs with only linear growth in the size of the program. 
The converse relationship holds, too:
\lparseNegfree{} programs can represent arbitrary ${\cal PB}$-programs without
increasing the size of the representation.
Let $r$ be a ${\cal PB}$-clause
\[
kW \lar k_1W_1,\ldots, k_mW_m ,\n(l_1 V_1), \ldots, \n(l_nV_n).
\]
We define $f(r)$ as follows. If there is $i$, $1\leq i\leq n$,
such that $l_i=0$, we set $f(r)=\ \ kW\lar kW$ (in fact any
tautology would do). Otherwise, we set
\[
f(r) = \ \ kW \lar k_1W_1,\ldots, k_mW_m, 0V_1 (l_1-1),\ldots,
0V_n(l_n-1).
\]
Given a ${\cal PB}$-program $P$, we define $f(P)=\{f(r)\colon r\in P\}$.

\begin{theorem}
\label{mca-ca}
Let $P$ be a ${\cal PB}$-program. A set of atoms $M$ is a stable
model of $P$ (as defined for ${\cal PB}$-programs) if and only if 
$M$ is an lparse-stable model of $f(P)$.
\end{theorem}
\begin{proof}
First, we observe that $P$ and $f(P)$ have the same models. Next, 
similarly as before, we have that the lparse-reduct $[f(P)]^M$
contains a clause 
\[
1\{a=1\} \lar k_1W_1,\ldots k_mW_m,0V_1,\ldots,0V_n 
\]
if and only if $P^M$ contains a clause of the form
\[
kW \lar k_1W_1,\ldots k_mW_m
\]
such that $a\in \as(W)$ and $M \models k_i W_i$ for all $1 \leq i \leq
m$.
Since in the clauses of the first type 
l-atoms $V_i$ are always true, as before, the results of the canonical 
computations from $P^M$ and $[f(P)]^M$ with respect to $M$ of $P$ 
coincide (we recall that both reducts are Horn ${\cal PB}$-programs). 
Using this observation one can complete the proof by reasoning as in 
the previous proof. 
\end{proof}

Theorems \ref{ca-mca} and \ref{mca-ca} establish the equivalence of
{\em \lparseNegfree{}} programs and ${\cal PB}$-programs with respect to
the stable model semantics. The translations $e$ and $f$ also preserve 
models. The equivalence between {\em \lparseNegfree{}} programs and 
${\cal PB}$-programs extends to supported models under the following
concept of supportedness for lparse-programs.

\begin{definition}
Let $P$ be a \lparseNegfree{} program. A set of atoms $M$ is a supported
model of $P$ if $M$ is a model of $P$ and if for every atom $a\in M$
there is an l-clause $A \lar B_1,\ldots, B_n$ in $P$ such that $a\in
\as(A)$ and $M\models B_i$, $1\leq i\leq n$.
\end{definition}

Indeed, we have the following two theorems (we only sketch a proof of
one of them; the proof of the other one is similar).

\begin{theorem}
Let $P$ be a \lparseNegfree{} program. A set $M$ is an
lparse-supported model of $P$ if and only if $M$ is a supported model
of $e(P)$, as defined for ${\cal PB}$-programs. 
\end{theorem}
\begin{proof}
Let us denote $Q=e(P)$. Let $M$ be an lparse-supported model of $P$. We 
will show that $M$ is a supported model of $Q$. By our earlier
observations, $P$ and $Q$ have the same models. Thus, $M$ is a model of 
$Q$. To complete the argument, we need to show that $M\subseteq \hs(Q(M
))$. Let $a\in M$. Since $M$ is an lparse-supported model of $P$, there 
is an l-clause $r\in P$ such that $r = A \leftarrow B_1,\ldots, B_n$, 
$a\in \as(A)$ and $M\models B_i$ for every $i$, $1\leq i\leq
n$. It follows that $a\in \as(\hd(e_1(r)))$ and that
$M\models\bd(e_1(r))$. Since $e_1(r)\in Q$, $e_1(r)\in Q(M)$. Thus,
$a\in \hs(Q(M))$. It follows that $M\subseteq \hs(Q(M))$ and so $M$ is a
supported model of $Q$.

Conversely, let us assume that $M$ is a supported model of $Q$. Then $M$
is a model of $Q$ and so $M$ is a model of $P$, as well. Let $a\in M$.
It follows that $a\in \hs(Q(M))$. Since each clause of the form
$e_2(r)$ ($r\in P$) is a constraint, there is an l-clause $r\in P$ such
that clause $e_1(r)\in Q$ such that $M\models \bd(e_1(r))$ and $a\in 
\hs(\hd(e_1(r)))$. Let $r = A \leftarrow B_1,\ldots, B_n$. It follows
that $a\in \as(A)$ and that $M\models B_i$, $1\leq i\leq n$. Thus,
$M$ is an lparse-supported model of $P$.
\end{proof}

\begin{theorem}
Let $P$ be a ${\cal PB}$-program. A set of atoms $M$ is a stable
model of $P$ (as defined for ${\cal PB}$-programs) if and only if
$M$ is an lparse-stable model of $f(P)$.
\end{theorem}

It follows from the results in this section that the translations 
$e$ and $f$ uniformly preserve basic semantics of \lparseNegfree{} and 
${\cal PB}$-programs, and allow us to view \lparseNegfree{} programs as 
${\cal PB}$-programs and {\em vice versa}.

We also note that this equivalence demonstrates that lparse
programs with the semantics of stable models as defined in \cite{nss99} 
can be viewed as a generalization of normal logic programming. It 
follows from Theorems \ref{lp-mca} and \ref{mca-ca} that the encoding 
of normal logic programs as lparse programs, defined as the 
composition of the translation $P\mapsto P^{pb}$ described in Section 
\ref{nlp-mca} (we note that the constraint $PB$ belongs to the class 
$\mathcal{PB}$) and the translation $f$, preserves the semantics of 
models, supported models and stable models (an alternative proof of this 
fact, restricted to the case of stable models was first given in 
\cite{sns02} and served as a motivation for the class of lparse
programs and its 
stable-model semantics). This result is important, as it is not at all 
evident that the reduct used in \cite{sns02}, leads to fixpoints
that
generalize the semantics of stable models as defined in \cite{gl88}.

Given that the formalisms of \lparseNegfree{} and ${\cal PB}$-programs are 
equivalent, it is important to stress what differentiates them. The 
advantage 
of the formalism of \lparseNegfree{} programs is that it does not require 
the negation operator in the language. The strength of the formalism of 
${\cal PB}$-programs lies in the fact that its syntax so closely 
resembles that of normal logic programs, and that the development of the 
theory of ${\cal PB}$-programs so closely follows that of the normal 
logic programming.

\section{Monotone $\cal F$-programs and disjunctive logic programs}
\label{mca-dp}

\cite{si94} introduced and investigated a semantics of {\em possible 
models} of disjunctive logic programs. It turns out that this semantics 
is different from the semantics proposed by Minker \cite{minker} and 
from that of Gelfond and Lifschitz \cite{gl90,Przymusinski91-NGC}. 
In this section, we 
will show that the formalism of monotone $\cal F$-programs generalizes the 
semantics of possible models. For the purpose of our discussion, we 
will extend the use of the terms {\em head}, {\em body}, 
$M$-applicability, and notation $P^M$, $\hd(r)$, $\bd(r)$ to the case 
of disjunctive programs.

\begin{definition}\cite{si94}
Let $P$ be a disjunctive logic program. A {\em split} program for $P$
is any normal logic program that can be obtained by the following
procedure. First, we select for each clause $r$ in $P$, a set $S_r
\subseteq \hd(r)$. Next, we replace $r$ with clauses of the form 
$A\leftarrow \body$, where $A\in S_r$, if $S_r\not= \emptyset$, and with 
the (constraint) clause $\ \leftarrow \bd(r)$, if $S_r=\emptyset$.

A set $M$ of atoms is a {\em possible} model of $P$ if $M$ is a stable 
model (in the sense of Gelfond and Lifschitz \cite{gl88})
of a split program for $P$. 
\ \ $\proofbox$
\end{definition}

We point out that we allow for disjunctive clauses to have empty heads 
(that is, we allow constraint clauses). We also allow that sets $S_r$ be 
empty. Consequently, split programs may contain constraints even if the
original program does not.

\begin{example}
Let $P$ be the disjunctive program:
\[
P = \{ a\vee b\vee c \leftarrow,\ \ a\vee c\leftarrow \n(b),\ \
b\leftarrow \n(c),\ \ c\leftarrow \n(a)\}.
\]
Then, the program $Q$:
\[
Q=\{ a \leftarrow,\ \ a\leftarrow \n(b), \ \ c\leftarrow \n(b),\ \
b\leftarrow \n(c),\ \ \leftarrow \n(a)\}.
\]
is an example of a split program for $P$ (given by the following subsets
of the sets of atoms in the heads of the clauses: $\{a\}$, $\{a,c\}$, $\{
b\}$ and $\emptyset$, respectively). Moreover, since $\{a,b\}$ is a 
stable model of $Q$, $\{a,b\}$ is a possible model of $P$.
\ \ $\proofbox$
\end{example}

If $M$ is a model of a disjunctive program $P$, by $s(P,M)$ we denote
the split program of $P$ determined by sets $S_r=M\cap \hd(r)$. We have
the following simple characterization of possible models.

\begin{proposition}
\label{pos-char}
Let $P$ be a disjunctive program. A set of atoms $M$ is a possible model
of $P$ if and only if $M$ is a stable model of the split program
$s(P,M)$.
\end{proposition}
\begin{proof}
If $M$ is a possible model of $P$, then $M$ is a stable model of a split
program of $P$, say $Q$. Let us assume that $Q$ is determined by sets 
$S_r\subseteq \hd(r)$, where $r\in P$. 

For every clause $r\in P$, if $r$ is $M$-applicable then all clauses it 
contributes to $Q$ are $M$-applicable, too, as they have the same body 
as $r$. Since $M$ is a model of $Q$ (being a stable model of $Q$), we 
obtain that $S_r \subseteq M$. Thus, $Q(M)\subseteq s(P,M)$.

Directly from the definition of $s(P,M)$ we obtain that $M$ is a model 
of $s(P,M)$. Thus, $M$ is a model of $[s(P,M)]^M$ and,
consequently, $\lm([s(P,M)]^M)$ exists. Moreover, it follows that
$\lm([s(P,M)]^M) \subseteq M$ (indeed, all non-constraint clauses in 
$s(P,M)$ have heads from $M$). 

Since $M$ is a stable model of $Q$, $M=\lm(Q^M)$. Thus, it follows that 
$M=\lm([Q(M)]^M)$ and so, we obtain:
\[
M= \lm([Q(M)]^M)\subseteq \lm([s(P,M)]^M)\subseteq M.
\]
Thus, $\lm([s(P,M)]^M) =M$ or, in other words, $M$ is a stable model of
$s(P,M)$. The converse implication follows by the definition. 
\end{proof}

Let $r$ be a disjunctive logic program clause of the form:
\[
c_1\vee\ldots\vee c_k \lar a_1,\ldots,a_m,\n(b_1),\ldots,\n(b_n),
\]
where all $a_i$, $b_i$ and $c_i$ are atoms. We encode this clause as a 
program clause with cardinality atoms:
\[
r^{ca}=\ \ 1\{c_1,\dots,c_k\} \lar 1\{a_1\},\ldots,1\{a_m\},
            \n(1\{b_1\}),\ldots,\n(1\{b_n\}).
\]
(If all $a_i$ and $b_i$ are distinct, the following translation could 
be used instead: $1\{c_1,\dots,c_k\} \lar m\{a_1,\ldots,a_m\}, 
\n(1\{b_1,\ldots,b_n\})$.) We note that if $k=0$, that is, the head of 
$r$ is empty, the rule $r^{ca}$ has the constraint $1\emptyset$ in the
head, which is inconsistent. In this case, $r^{ca}$ is a constraint
clause. 

For a disjunctive logic program $P$, we define $P^{ca}=\{r^{ca} \colon
r\in P\}$ ($ca$ in the subscript stands for ``cardinality atoms'').
Since cardinality constraints are monotone, the concept of a stable
model of the program $P^{ca}$ is well defined. We have the following 
theorem.

\begin{theorem}
Let $P$ be a disjunctive logic program. A set of atoms $M$ is a
possible model of $P$ if and only if $M$ is a stable model of the
program $P^{ca}$ (in the sense, we defined in this paper).
\end{theorem}
\begin{proof} 
We first note that $[s(P,M)]^M =s(P^M,M)$. Thus, by Proposition 
\ref{pos-char}, it follows that $M$ is a possible model of $P$ if and
only if $M$ is a least model of $s(P^M,M)$. We also note that $[P^{ca}
]^M=[P^M]^{ca}$. Thus, $M$ is a stable model of $P^{ca}$ if and 
only if $M$ is a derivable model of $[P^M]^{ca}$. 

It follows that in order to prove the assertion it suffices to show that
for every positive (no negation in the bodies of clauses) disjunctive
program $P$, $M$ is a least model of $s(P,M)$ if and only if $M$ is
a derivable model of $P^{ca}$. We will now prove this claim. To simplify
notation, we will write $Q$ instead of $P^{ca}$. 

First, we note $P$ and $Q$ have the same models. Thus, each side of the 
equivalence implies that $M$ is a model of $Q$. In particular, it
follows (no matter which implication we are proving) that $Q$ has a 
canonical computation $t^{Q,M} = (X^{Q,M}_n)_{n=0, 1,\ldots}$. Next, 
we observe that for every $X\subseteq M$, the definitions of $Q$ and 
$s(P,M)$ imply that
\[
\hs(Q(X)) \cap M = \hd(s(P,M)(X)) = T_{s(P,M)}(X).
\]
In particular, since $X^{Q,M}_{n+1} = \hs(Q(X^{Q,M}_n))\cap M$, for every 
$n=0,1,\ldots$, we have
\[
X^{Q,M}_{n+1} = T_{s(P,M)}(X^{Q,M}_n).
\]
These identities imply that the result of the canonical $Q$-computation 
for $M$ and the least model of $s(P,M)$ coincide. Consequently, $M$ is 
a derivable model of $Q$ if and only if $M$ is a least model of 
$s(P,M)$ as claimed. 
\end{proof}

\section{Discussion}\label{disc}

In the paper, we introduced and studied the formalism of $\cal 
F$-programs. When all constraints in $\cal F$ are monotone, this 
formalism offers an abstract framework for integrating constraints into 
logic programming. It exploits and builds on analogies 
with normal logic programming. Most concepts and techniques for monotone
$\cal F$-programs are closely patterned after their counterparts developed 
there and so, normal logic programming can be viewed as a fragment of 
the theory of monotone $\cal F$-programs. Importantly, the same is the 
case for other nonmonotonic systems namely, the disjunctive logic 
programming with the possible-model semantics of \cite{si94}, and for 
the formalism of logic programs with weight constraints \cite{sns02}. 
For these two formalisms, monotone $\cal F$-programs help to explain 
the nature of their 
relationship with normal logic programming, hidden by their original 
definitions.

In this paper, we developed a sound foundation for the theory of monotone
$\cal 
F$-programs. Recently, the theory of monotone $\cal F$-programs was 
developed
further. \cite{lt05} demonstrated that Fages lemma \cite{fag94}, and the
concepts of the program completion and a loop formula extend to the
setting of monotone $\cal F$-programs. The latter two properties allow one to 
reduce stable-model computation for programs with weight constraints
to the problem of computing models of propositional theories extended
with weight atoms (referred to as {\em pseudo-boolean} constraints in 
the satisfiability community). \cite{lt05b} exploited this reduction
and developed an algorithm to compute stable models of programs with
weight constraints by using off-the-shelf solvers of pseudo-boolean 
constraints such as those described in
\cite{bar95,wal97,arms02,lt03,et04a}. 

There are strong analogies between the approach we propose and develop
here and some of the {\em techniques} discussed in \cite{si94} in the
context of disjunctive programs with the semantics of possible models.
One way to look at the results of our paper is that it extends the way 
\cite{si94} handles nondeterminism, inherent in disjunctive logic 
programs, to the abstract setting of monotone $\cal F$-programs. In 
particular, 
\cite{si94} presents a computational procedure for disjunctive programs 
without negation, which can be shown to be closely related to our notion 
of a $P$-computation. That paper also introduces a nondeterministic
provability operator, defined for {\em positive} disjunctive programs. 
Three aspects differentiate our work from \cite{si94}. Most importantly, 
we study here a much broader class of programs than disjunctive ones. 
Secondly, we define a provability operator on the class of {\em all} 
monotone 
$\cal F$-programs and not just positive ones. Finally, we consistently 
exploit properties of this operator, and align our approach with the 
standard operator-based development of normal logic programming 
\cite{ap90,fi99}.

The emergence of a nondeterministic one-step provability operator warrants
additional comments. Nondeterministic provability operators were considered 
before in the context of logic programming. We already noted that 
\cite{si94} proposed a provability operator similar to the one we 
introduced here (although only for the class of {\em positive} disjunctive 
programs). \cite{si95a} proposed another operator designed to capture a 
different computational process arising in the context of paraconsistent 
systems. Finally, \cite{pt04} presented a characterization of answer sets 
of disjunctive logic programs in terms of yet another nondeterministic 
provability operator. However, the operator we proposed here exhibits the 
closest parallels with the van Emden-Kowalski operator and opens up a 
possibility of generalizing the approximation theory proposed in 
\cite{dmt00a} to the case of monotone $\cal F$-programs. However, for 
that to 
happen, one will need techniques for handling nondeterministic operators 
on lattices, similar to those presented for the deterministic operators 
in \cite{dmt00a,dmt02}. Developing such techniques is an open problem.

\section*{Acknowledgments}

The authors wish to thank the anonymous reviewers for their detailed 
comments and suggestions, which helped improve the paper. The second 
author was supported by the Academy of Finland grant 211025. The other 
two authors were supported by the NSF grants IIS-0097278 and IIS-0325063.

\pagerange{\pageref{firstpage}--\pageref{lastpage}}

\bibliographystyle{acmtrans}

\end{document}